\documentclass{article}

\usepackage{arxiv}
\usepackage[utf8]{inputenc} 
\usepackage[T1]{fontenc}    
\usepackage{hyperref}       
\usepackage{url}            
\usepackage{booktabs}       
\usepackage{amsfonts}       
\usepackage{nicefrac}       
\usepackage{microtype}      
\usepackage{lipsum}
\usepackage{graphicx}

\usepackage{makeidx}  
\usepackage{svg}
\usepackage{float}
\usepackage[export]{adjustbox}
\usepackage{verbatim}
\usepackage{amsmath}

\DeclareMathOperator*{\argmin}{arg\,min}
\usepackage{isomath}
\usepackage{bm}
\usepackage{fancyhdr,graphicx,amsmath,amssymb}
\usepackage[ruled,vlined]{algorithm2e}
\usepackage{algorithmic}
\include{pythonlisting}
\usepackage{times}
\usepackage{amssymb}
\usepackage{setspace}
\usepackage{color}
\usepackage{multirow}
\usepackage{url}

\newcommand{\vect}[1]{\bm{#1}}
\newcommand{\matr}[1]{\mathbf{#1}}

\usepackage{subfigure}

\graphicspath{ {./figures/} }

\title{Stereo Camera Visual SLAM with Hierarchical Masking and Motion-state Classification at Outdoor Construction Sites Containing Large Dynamic Objects}

\author{
 Runqiu Bao \\
  Dept. of Precision Engineering\\
  The University of Tokyo, Tokyo, Japan \\
  \texttt{bao@robot.t.u-tokyo.ac.jp} \\
  \And
 Ren Komatsu \\
  Dept. of Precision Engineering\\
  The University of Tokyo, Tokyo, Japan \\
  \texttt{komatsu@robot.t.u-tokyo.ac.jp} \\
  \And
 Renato Miyagusuku \\
  Dept. of Mechanical and Intelligent Engineering\\
  Utsunomiya University, Utsunomiya, Tochigi, Japan \\
  \texttt{miyagusuku@cc.utsunomiya-u.ac.jp} \\
  \AND
  Masaki Chino \\
  Construction Division \\
  HAZAMA ANDO CORPORATION, Tokyo, Japan \\
  \texttt{chino.masaki@ad-hzm.co.jp} \\
  \And
  Atsushi Yamashita \\
  Dept. of Precision Engineering\\
  The University of Tokyo, Tokyo, Japan \\
  \texttt{yamashita@robot.t.u-tokyo.ac.jp}
  \And
  Hajime Asama \\
  Dept. of Precision Engineering\\
  The University of Tokyo, Tokyo, Japan \\
  \texttt{asama@robot.t.u-tokyo.ac.jp}
}

\begin{document}

\thanks{Code available at: https://github.com/RunqiuBao/kenki-positioning-vSLAM}
\thanks{$^\ast$Corresponding author Email: bao@robot.t.u-tokyo.ac.jp\vspace{6pt}}

\maketitle
\begin{abstract}
At modern construction sites, utilizing GNSS (Global Navigation Satellite System) to measure the real-time location and orientation (i.e. pose) of construction machines and navigate them is very common. However, GNSS is not always available. Replacing GNSS with on-board cameras and visual simultaneous localization and mapping (visual SLAM) to navigate the machines is a cost-effective solution. Nevertheless, at construction sites, multiple construction machines will usually work together and side-by-side, causing large dynamic occlusions in the cameras' view. Standard visual SLAM cannot handle large dynamic occlusions well. In this work, we propose a motion segmentation method to efficiently extract static parts from crowded dynamic scenes to enable robust tracking of camera ego-motion. Our method utilizes semantic information combined with object-level geometric constraints to quickly detect the static parts of the scene. Then, we perform a two-step coarse-to-fine ego-motion tracking with reference to the static parts. This leads to a novel dynamic visual SLAM formation. We test our proposals through a real implementation based on ORB-SLAM2, and datasets we collected from real construction sites. The results show that when standard visual SLAM fails, our method can still retain accurate camera ego-motion tracking in real-time. Comparing to state-of-the-art dynamic visual SLAM methods, ours shows outstanding efficiency and competitive result trajectory accuracy.\medskip

\medskip
\end{abstract}


\section{Introduction}

\par Knowledge of real-time location and orientation (i.e. pose) of construction machines, such as bulldozers, excavators, and vibration rollers, is essential for the automation of construction sites. Currently, RTK-GNSS (Real-Time Kinematic - Global Navigation Satellite System) is widely used because of its centimeter-level location accuracy. However, in addition to the high price, the location output of RTK-GNSS could be unstable due to loss of satellite signals underground, near mountains and trees, and between tall buildings. Therefore, replacing RTK-GNSS with on-board cameras and visual SLAM (vSLAM) has been proposed \cite{gccebao}. Assuming machine's starting pose is known in a global coordinate system, relative pose outputs from vSLAM can be used to navigate the machine.
\par However at construction sites, several machines usually work together and side-by-side (Figure~1), which results in large dynamic occlusions in the view of the cameras. Such dynamic occlusions can occupy more than 50\% of the image. It leads to a dramatic drop in tracking accuracy or even tracking failure when using standard vSLAM. We introduce this problem distinctly in the context of dynamic vSLAM and propose an original stereo camera dynamic vSLAM formation.
\par To deal with dynamic occlusions, our idea is to firstly detect static objects and backgrounds, and then track ego-motion with reference to them. To achieve this, we need to estimate the real motion-states of objects. We use learning-based object detection and instance segmentation combined with object-wise geometric measurement in stereo frames to label the motion-states of object instances and generate occlusion masks for dynamic objects. Additionally, two types of occlusion masks are applied to balance accuracy and computation cost, bounding box mask for small occlusions and pixel-wise for large occlusions. Pixel-wise masks describe boundaries of objects more accurately. While bounding boxes are faster to predict, it is not so accurate. 
\par In a nutshell, our contributions in this work include: (1) a semantic-geometric approach to detect static objects and static backgrounds for stereo vSLAM, (2) a masking technique for dynamic objects called hierarchical masking, (3) a novel stereo camera dynamic visual SLAM system for construction sites.
\par The remainder of this work is organized as follows: In Section~2, we summarize the existing research on dynamic visual SLAM and motion segmentation methods, and describe the features of this work. In Section~3, the system structure and our original proposals (two algorithms) are introduced. In Section~4, to test the performance of our proposals, we conducted experiments at real construction sites and built datasets for algorithm evaluation. We used Absolute Trajectroy RMSE \cite{bescos2018dynaslam} to evaluate accuracy of the location outputs of the vSLAM system. Finally, Section~5 contains the conclusions and future work plan.

\begin{figure}[b!]
    \centering
    \includegraphics[scale=0.38]{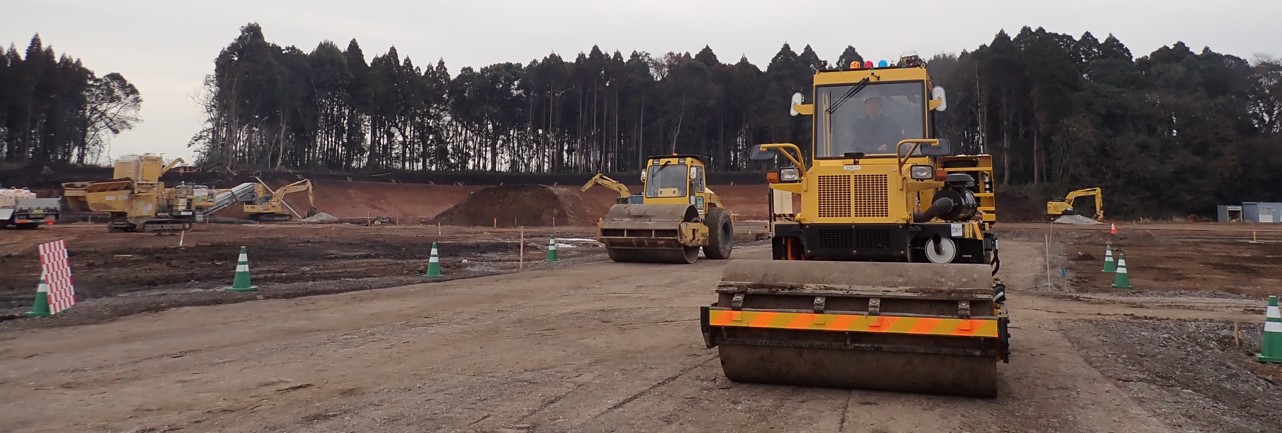}
    \caption{Simultaneous working of construction machines causing large-area moving occlusions in on-board cameras'  view.}
\end{figure}

\section{Related Work}
\subsection{Dynamic Visual SLAM}
Standard visual SLAM (vSLAM) assumes that the environment is static. Correspondingly, vSLAM for dynamic environments (Dynamic vSLAM or Robust vSLAM) distinguishes static and dynamic features and computes pose estimation based solely on static features. 
\par Depending on the application, dynamic vSLAM can be categorized into two classes. One solely builds a static background model, ignoring moving objects \cite{jaimez2017icra, barnes2018icra, bescos2018dynaslam}. The other aims at not only creating a static background map, but simultaneously maintaining sub-maps of moving objects \cite{xu2019mid, barsan2018icra, runz2018maskfusion}. Our task, i.e. positioning of construction machines, requires fast and accurate camera ego-motion tracking and thus belongs to the first class. 

\par Real-time positioning task at construction sites brought new problem to vSLAM. Specifically, we found that at a busy construction site, there are often many machines, trucks and persons moving around which become large dynamic occlusions (occlusion rate \textgreater{50\%} from time to time) in the camera view. Besides, such occlusions usually contain more salient feature points than earthen ground and cause chaos in feature-based camera ego-motion tracking. Even existing dynamic vSLAM solutions may suffer from different issues and are thus not the optimal solution to this task. For example, \cite{ds-slam2018, detect-slam2018, dynamic-slam2019, soares2019visual} proposed very fast methods for dealing with dynamic objects. Yet, they did not explicitly consider overly-large dynamic occlusions and thus might suffer from accuracy drop. \cite{bescos2018dynaslam} and \cite{barsan2018icra} proposed very robust methods for masking dynamic occlusions. But both of them require heavy computation and are not suitable for real-time positioning task. Therefore, we proposed our own dynamic vSLAM solution for real-time positioning at dynamic construction sites.

\par In a dynamic vSLAM system, there are mainly two major modules: (1) motion segmentation and (2) localization and mapping \cite{saputra2018survey}. Motion segmentation is the key part that distinguishes an outstanding dynamic vSLAM system from the rests. 

\begin{figure}[t]
    \centering
    \includegraphics[scale=0.4]{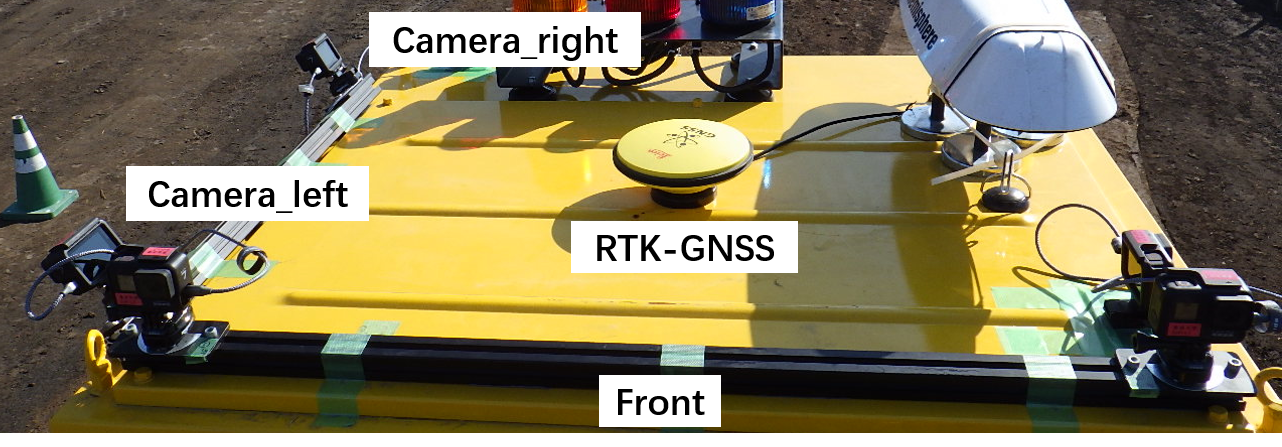}
    \caption{Cameras are mounted on top of our construction machine facing to the sides, and RTK-GNSS is used to collect ground truth positions.}
\end{figure}

\subsection{Motion Segmentation}
Motion segmentation is aimed at detecting moving parts in the image and classifying the features into two groups, static and dynamic features.
\par Standard visual SLAM achieves this by applying robust statistical approaches to the estimation of geometric models, such as Random Sample Consensus (RANSAC) \cite{fischler1981ransac}. However, such approach may fail when large dynamic occlusions exist, and static features are not in the majority. Other approaches leverage external sensors such as inertial measurement units (IMU) to fix camera ego-motion. In the following, we focus on visual-only approaches to distinguish static and dynamic features. Muhamad et al. \cite{saputra2018survey} summarizes this research area well, for more details please refer to the study.
\par The most intuitive approach for motion segmentation is using semantic information to separate object instances that may move in the scene. To obtain semantic information, B{\^a}rsan et al. \cite{barsan2018icra} used learning-based instance segmentation to generate pixel-wise masks for object instances. Cui et al. \cite{cui2019icra} proposed only using bounding boxes obtained from YOLO v3 \cite{redmon2018yolov3} to filter dynamic objects, which can reduce computation cost. However, these works simply assume that movable objects are dynamic. End-to-end learning-based methods for motion segmentation (without prior information about the environment) are still scarce \cite{saputra2018survey}.
\par Another common strategy for motion segmentation is utilizing geometric constraints. It leverages the fact that dynamic features will violate constraints defined in multi-view geometry for static scenes. Kundu et al. \cite{kundu2009moving} detected dynamic features by checking if the points lie on the epipolar line in the subsequent view and used Flow Vector Bound (FVB) to distinguish motion-states of 3D points moving along the epipolar line. Migliore et al. \cite{migliore2009icra} kept checking the intersection between three projected viewing rays in three different views to confirm static points. Tan et al. \cite{tan2013robust} projected existing map points into the current frame to check if the feature is dynamic. It is difficult for us to evaluate these methods. However, one obvious drawback is that they require complicated modifications to the bottom components of standard visual SLAM algorithm without the static environment assumption. We argue that such modifications are not good for the modularity of a vSLAM system.
\par As a novel hybrid approach, Berta et al. \cite{bescos2018dynaslam}, in their work named Dynaslam, proposed to combine learning-based instance segmentation with multi-view geometry to refine masks for objects that are not a priori dynamic, but movable. Our system follows the hybrid fashion of Dynaslam, but we treat motion segmentation as an object-level classification problem. Our idea is, by triangulating and measuring positions of points inside the bounding boxes and comparing them between frames, we can estimate object-level motion-states for every bounding box (assuming objects are all rigid). If we know the motion-state of every bounding box, the surroundings can be easily divided into static and dynamic parts. 

Besides, bounding boxes of large dynamic occlusions reduce available static features. We will show that it is essential to keep the overall masked area under a certain threshold if possible. Hence, we designed an algorithm named hierarchical masking to refine a pixel-wise mask inside the bounding box when the overall masked area extends past a threshold to save scarce static features. This hierarchical masking
algorithm is also an original proposal from us.

\section{Stereo Camera Dynamic Visual SLAM robust against Large Dynamic Occlusions}

The core problem in this research is to achieve fast and accurate camera ego-motion tracking when there are large occlusions in the camera’s view. Subsection~3.1 is a general introduction of the system pipeline. In Subsection~3.2, the principle of feature-based camera ego-motion tracking with occlusion masks for dynamic occlusions is introduced. In order to balance computation speed and accuracy in occlusion mask generation, a hierarchical masking approach is proposed in Subsection~3.3. Last, through stereo triangulation and comparison, object instances in the current frame will be assigned a predicted motion-state label, static or dynamic, which leads to further mask refining and a second around of tracking.

\begin{figure}[t]
    \centering
    \includegraphics[width=6cm]{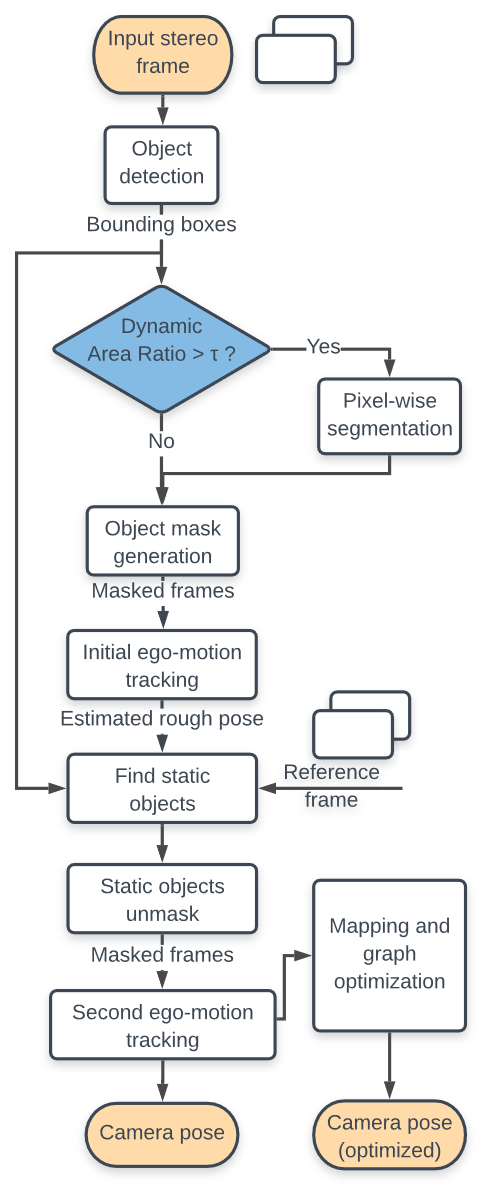}
    \caption{An overview of the proposed system. Inputs are stereo frames (all the processes are on the left image. Right image is only for triangulating 3D points). After semantic information extraction, occlusion masks of objects are generated and used in filtering potential dynamic features. The initial ego-motion tracking is based purely on the static background. Then more static objects are found and used as references in the second round of tracking to get more accuracy results. The final output is the camera pose $\matr{R}$ and $\vect{t}$ of the current frame in the SLAM coordinates.}
\end{figure}

\subsection{System Overview}

The system installation is illustrated in Figure~2 and the system pipeline is shown in Figure~3. Inputs are stereo frames (left image and right image) captured by a stereo camera. Then semantic information, including object labels and bounding boxes, are extracted using learning-based object detection. In addition, a hierarchical mask generation approach is proposed to balance mask accuracy and generation speed. Object masks exclude suspicious dynamic objects from the static background. The features in the static background are then used in the initial tracking of camera pose.
\par After initial tracking, a rough pose of the new frame is known, with which we distinguish static objects from other objects. This is done by triangulating object-level 3D key points in the reference and current frame and comparing the 3D position errors to distinguish whether the object is moving or not. Large static objects can provide more salient static features for improving tracking accuracy. Dynamic objects will be kept masked in the second ego-motion tracking. This two-round coarse-to-fine tracking scheme helps detect static objects and improve pose estimation accuracy.
\par After the second round of tracking, there will be mapping and pose graph optimization steps as most of state-of-the-art vSLAM algorithms do.

\subsection{Feature-based Camera Ego-motion Tracking by Masking Dynamic Occlusions}
\par Camera ego-motion tracking framework used here is based on ORB-SLAM2 stereo \cite{mur2017orbslam2}. When a new frame comes in, first, a constant velocity motion model is used to predict the new camera pose with which we can search for map points and 2D feature points matches. After enough matches are found, a new pose can be re-estimated by Perspective-n-point (PnP) algorithm \cite{mur2015orbslam}. Motion-only bundle adjustment (BA) is then used for further pose optimization. Motion-only BA estimates the camera pose of the new stereo frame, including orientation $\matr{R}\in SO\left(3\right)$ and position ${\vect{t}}\in \mathbb{R}^{3}$, by minimizing the reprojection error between matched 3D points $\vect{x_{i}}\in \mathbb{R}^{3}$ in the SLAM coordinates and feature points $\vect{p^{i}_{\left(.\right)}}$ in the new frame, where $i=1,2,...,N$. $\vect{p^{i}_{\left(.\right)}}$ include monocular feature points $\vect{p^{i}_{m}}\in \mathbb{R}^{2}$ and stereo feature points $\vect{p^{i}_{s}}\in \mathbb{R}^{3}$.
\par Now supposing $M$ out of $N$ 3D points are on a rigid body dynamic object that had a pose change $\matr{R'}, \vect{t'}$ in the physical world, and their 3D coordinates change from $\vect{x_{i}}$ to $\vect{x_{i}'}$, for $i=1,2,...,M$. The rigid body transformation can be expressed as $\vect{x_{i}'}=\matr{R'}\vect{x_{i}}+\vect{t'}$. Pose change estimation can be expressed as:
\begin{normalsize}
\begin{equation}
\begin{aligned}
\left\{ \matr{R},\vect{t},\matr{R'},\vect{t'}\right\}=
&  \argmin_{\matr{R},\vect{t},\matr{R'},\vect{t'}} \bigg [ \sum ^{M}_{i=1}\rho\left( \Big \| \vect{p^{i}_{\left(.\right)}}-\pi_{\left(.\right)}\left( \matr{R}\left(\matr{R'}\vect{x_{i}'+\vect{t'}}\right)+\vect{t}\right) \Big \|^{2}_{\matr{\Sigma}}\right)\\
&+\sum^{N}_{M+1}\rho \left(\Big \|\vect{p^{i}_{\left(.\right)}}-\pi_{\left(.\right)}\left( \matr{R}\vect{x_{i}}+\vect{t}\right) \Big \|^{2}_{\matr{\Sigma}}\right) \bigg ],
\end{aligned}
\end{equation}
\end{normalsize}\\
where $\rho$ is the robust Huber cost function that controls the error growth of the quadratic function, and $\matr{\Sigma}$ is the covariance matrix associated to the scale of the feature point. The project functions $\pi_{\left(.\right)}$ include monocular $\pi_{\mathrm{m}}$ and rectified stereo $\pi_{\mathrm{s}}$, as defined in \cite{mur2017orbslam2}:
\begin{normalsize}
\begin{equation}
\begin{aligned}
\pi _{\mathrm{m}}\left( \begin{bmatrix} X \\ Y \\ Z \end{bmatrix}\right) =\begin{bmatrix} f_{\mathrm{x}}X/Z+c_{\mathrm{x}} \\ f_{\mathrm{y}}Y/Z+c_{\mathrm{y}} \end{bmatrix}=\begin{bmatrix}
u_{\mathrm{l}} \\
v_{\mathrm{l}} \\
\end{bmatrix},
\end{aligned}
\end{equation}
\end{normalsize}

\begin{normalsize}
\begin{equation}
\begin{aligned}
\pi _{\mathrm{s}}\left( \begin{bmatrix} X \\ Y \\ Z \end{bmatrix}\right) =\begin{bmatrix} f_{\mathrm{x}}X/Z+c_{\mathrm{x}} \\ f_{\mathrm{y}}Y/Z+c_{\mathrm{y}} \\ 
f_{\mathrm{x}}\left(X-b\right)/Z+c_{\mathrm{x}} \end{bmatrix}=\begin{bmatrix}
u_{\mathrm{l}} \\
v_{\mathrm{l}} \\
u_{\mathrm{r}}
\end{bmatrix},
\end{aligned}
\end{equation}
\end{normalsize}\\

\noindent where $\left(f_{\mathrm{x}}, f_{\mathrm{y}}\right)$ is the focal length, $\left(c_{\mathrm{x}}, c_{\mathrm{y}}\right)$ is the principal point and b the baseline. $\left(u_{\mathrm{l}}, v_{\mathrm{l}}\right)$ represents the monocular feature points and $\left(u_{\mathrm{l}}, v_{\mathrm{l}}, u_{\mathrm{r}}\right)$ the stereo feature points.  
\par However, solving this equation~(1) is not easy, not to mention that there could be more than one dynamic objects in real world. If we only want to estimate $\matr{R}, \vect{t}$, equation~(1) can be simplified to:

\begin{normalsize}
\begin{equation}
\begin{aligned}
\left\{ \matr{R},\vect{t}\right\} = \argmin_{\matr{R},\vect{t}}\sum ^{N}_{i=M+1}\rho\left( \bigg \| \vect{p^{i}_{\left(.\right)}}-\pi_{\left(.\right)}\left( \matr{R}\vect{x_{i}}+\vect{t}\right) \bigg \|^{2}_{\matr{\Sigma}} \right),
\end{aligned}
\end{equation}
\end{normalsize}\\

\noindent which means only using static points in the scene to estimate the camera pose. If dynamic feature points as moving outliers are not excluded, the estimation result will be wrong. 
\par To separate static and dynamic feature points, our approach is to use a binary image as mask (for the left image of the input stereo frame). The mask has the same size as the input image, and pixels with value 0 indicate static area, while pixels with value 1 indicate dynamic area. Suppose that $\rm I_{mask}\left(u,v\right)$ refers to a pixel in the mask image $\rm I_{mask}$. $S_p$ is a set of static pixels and $D_p$ is a set of dynamic pixels,
\begin{normalsize}
\begin{equation}
{\rm I_{mask}}\left( u,v\right) =\begin{cases}0, \quad {\rm I_{mask}}\left( u,v\right) \in S_p\\
1, \quad {\rm I_{mask}}\left( u,v\right) \in D_p\end{cases}.
\end{equation}
\end{normalsize}\\

Figure~4 shows examples of mask (with alpha blending). To generate a mask, we first get bounding boxes or pixel-wise segmentation results from learning-based object detection and instance segmentation (Subsection~3.3). Then, for those objects with a priori dynamic semantic label such as "car", "person", "truck", etc., we change the pixels' value to 1 in the mask image, while keeping the others as 0. We also apply geometrical measurement and calculate a motion-state label for every object bounding box. Inside a static bounding box, we change the pixels' value to 0 whatever it was (Subsection 3.4). Later during ego-motion tracking period, only the areas where the mask value equals 0 will be used to estimate camera pose as described by Equation~(4).

\subsection{Hierarchical Object Masking}

The switching between two types of masks forms a hierarchical masking strategy that balances computation speed and mask accuracy.
\par To reduce computation cost, we first used object detectors, e.p. EfficientDet \cite{tan2019efficientdet}, to predict object instances and recognize their bounding boxes. Such learning-based object detector is a deep neural network, which can predict all the bounding boxes, class labels, and class probabilities directly from an image in one evaluation. A bounding box only represents a rough boundary of the object, so when using it as an object mask, background feature points inside the rectangle are also classified as "object". It is, therefore, only a rough boundary description.
\par There were cases when bounding boxes occupied most of the area in the image, which led to a shortage of available static features, and thus the accuracy of the ego-motion tracking declined. In such cases, we perform pixel-wise segmentation on the image to save more static features. For pixel-wise segmentation, we also use deep learning approaches, such as Mask R-CNN \cite{he2017maskrcnn}. Pixel-wise segmentation takes more time and slows down the system output rate. Thus, only in extreme cases when the frame is so crowded with object bounding boxes, should pixel-wise segmentation be performed.
\par The switching to pixel-wise segmentation is controlled by an index named Masked Area Ratio ($mar$). If $A_{m}$ is the total area of bounding boxes in pixels and $A_{f}$ is the total area of the image in pixels, then we have,
\begin{normalsize}
\begin{equation}
\begin{aligned}
mar=
\dfrac{A_{m}}{A_{f}}.
\end{aligned}
\end{equation}
\end{normalsize}
\par If $mar$ is larger than the threshold $\tau_{\rm mar}$, it means the current frame is quite crowded and pixel-wise segmentation is necessary.

\par Hierarchical object masking is concluded as following: when we get one frame input, we first use object detector performing object detection and obtain bounding boxes. Then $mar$ is calculated. If $mar$ is higher than a pre-set threshold $\tau_{\rm mar}$, then we perform pixel-wise segmentation and output the pixel-wise object mask. If $mar$ is smaller than the threshold, then the bounding box mask are directly forwarded as object mask. This algorithm is summarized in Algorithm~1.

\begin{figure}[t!]
    \centering
    \includegraphics[scale=0.5]{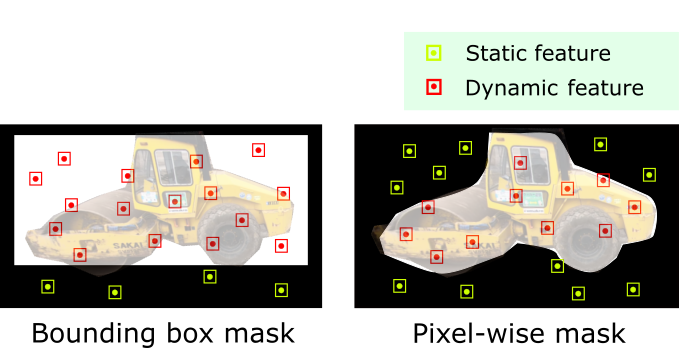}
    \caption{Two kinds of masks and masked features.}
\end{figure}

\begin{algorithm}[t!]
\caption{Hierarchical Masking}
\begin{algorithmic}[1]
\renewcommand{\algorithmicrequire}{\textbf{Input:}}
\renewcommand{\algorithmicensure}{\textbf{Output:}}
\REQUIRE stereo images in current frame, $\rm I_{cl}$, $\rm I_{cr}$; Mased Area Ratio threshold, $\tau_{\rm mar}$.
\ENSURE  image mask for the left image in current frame, $\rm I_{mask}$.
\\ \textit{Initialisation}: a blank image mask, $\rm I_{mask}$; initial Masked Area Ratio as 0, $\rm mar = 0$\;
\STATE $\rm I_{mask}$=objectDetectionAndMasking($\rm I_{cl}$)\;
\STATE $\rm mar$=calMaskedAreaRatio($\rm I_{mask}$)\;
\IF {($\rm mar$ $\geq \tau_{\rm mar}$)}
\STATE $\rm I_{mask}$=pixelwiseSegmentationAndMasking($\rm I_{cl}$)\;
\ENDIF
\RETURN $\rm I_{mask}$
\end{algorithmic} 
\end{algorithm}

\begin{figure}[t!]
    \centering
    \includegraphics[scale=0.5]{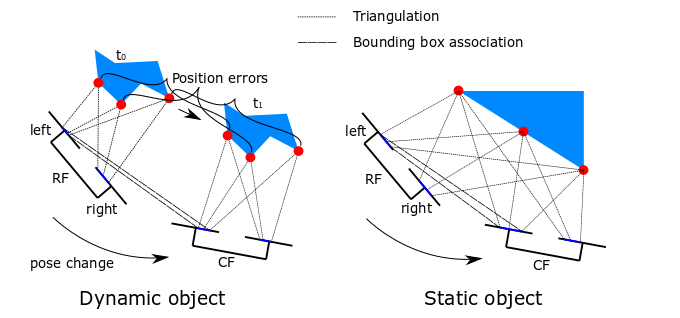}
    \caption{Associate bounding boxes between the Reference Frame (RF) and Current Frame (CF) using feature matching. Triangulate object-level 3D points in RF, then triangulate corresponding 3D points in CF and compare their positions in the two measurements. If most of point-wise position errors of an object (bounding box) are smaller than three times the standard variation of static background points, the object is labeled as `static' during camera pose change from RF to CF.}
\end{figure}

\begin{figure}[t!]
    \centering
    \includegraphics[scale=0.7]{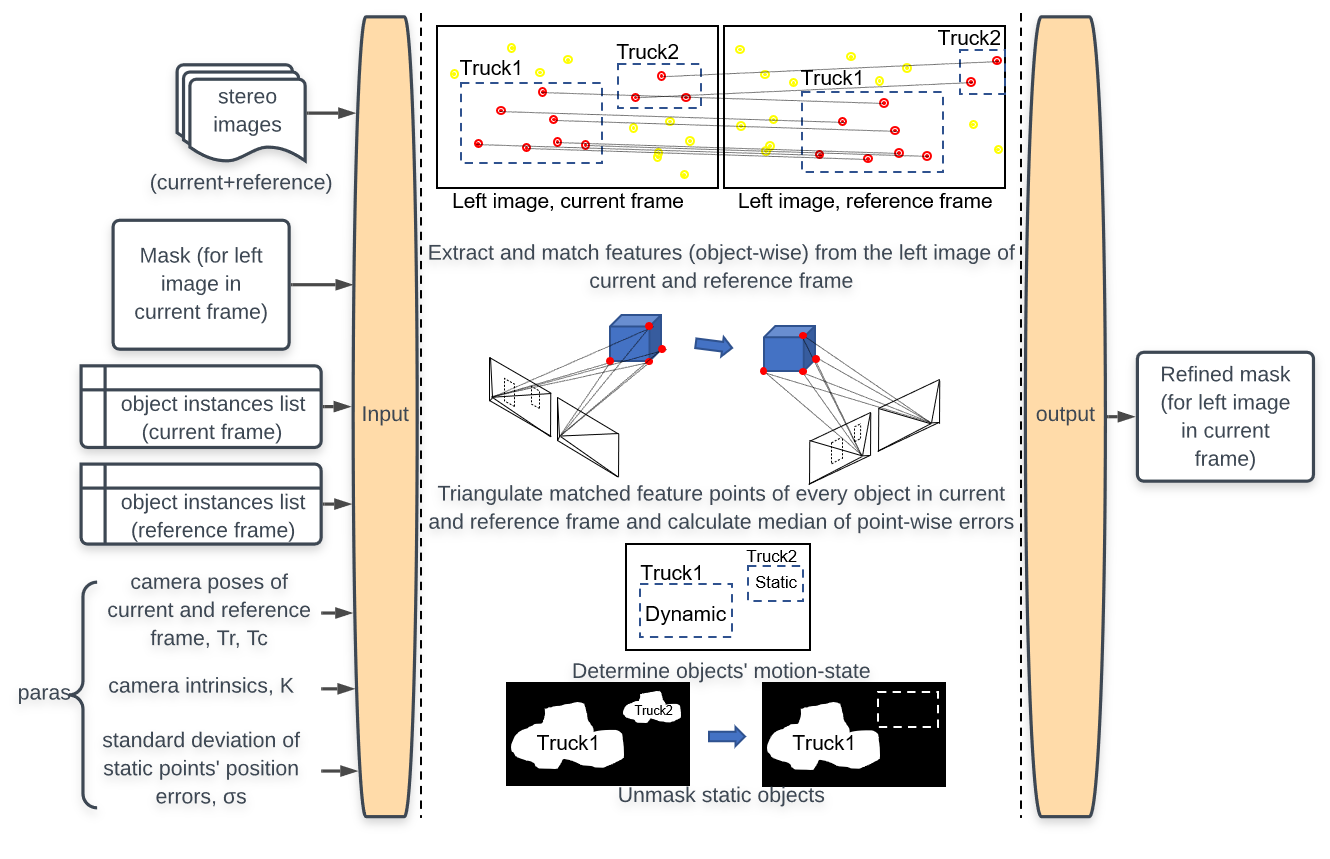}
    \caption{Algorithm 2: Objects' Motion-state Classification.}
\end{figure}

\subsection{\textbf{Objects' Motion-state Classification for Further Mask Refinement}}

\par After the first ego-motion tracking, with reference to the background, we roughly know the pose of the current frame. Based on the current pose, we triangulate object-level 3D points on all the detected object instances in the current frame and a selected reference frame and distinguish whether they have moved. Feature points inside static bounding boxes are then unmasked and used as valid static references in the second round of tracking. This algorithm (Algorithm 2) named motion-state classification is detailed in the following.

\par To classify objects' motion-state, first, a reference frame needs to be selected from previous frames. In this work, we used the $N$-th frame before the current frame as reference frame. $N$ is determined based on the machines' velocity. For example, for vibration rollers moving at 4 km/h mostly, $\rm FPS/3$ to $\rm FPS/2$ can be selected as $N$ ($\rm FPS$ stands for the frame rate of camera recording, namely Frame Per Second). For domestic automobiles running at higher speed, $N$ should be selected smaller so that there is appropriate visual change between current and reference frame. This strategy is simple but effective, given the simple moving pattern of construction machines. There are more sophisticated methods for selecting the best reference frame as stated in \cite{bescos2018dynaslam} and \cite{tan2013robust}.

\par Then, suppose there are objects $\{obj_{1},obj_{2},...,obj_{m}\}$ in the reference frame (RF) and objects $\{obj_{1},obj_{2},...,obj_{n}\}$ in the current frame (CF). We associate the m objects in RF with the n objects in CF by feature matching. If the object instances are associated successfully between two frames, which means the object is co-visible in both frames, we triangulate 3D points within the bounding boxes in both frames in SLAM coordinates and calculate point-wise position errors. 3D points' position errors of static background are assumed to obey zero-mean Gaussian distribution. The standard deviation, $\sigma_{bkg}$, is determined beforehand and used as the threshold for classification. For static objects, principally all 3D points' position error should be less than three times of $\sigma_{bkg}$. But considering the inaccuracy of a bounding box, we loosened the condition to $70\%$, i.e. objects are classified as "static" when more than 70\% of its 3D points have a position error smaller than $\left(3\times \sigma_{bkg}\right)$. However, outliers of feature matching usually result in very large position errors. We only keep points with position error smaller than the median to exclude outliers. Figure~5 shows the principle of the geometric constraint, the left one is a dynamic object and the right one is a static object. Figure~6 shows input and output as well as main ideas of Algorithm 2. Details about how to implement this algorithm can be found in our code repository.

\begin{figure}
\begin{center}
\subfigure[\label{a}Construction site bird view]{
\resizebox*{6cm}{!}{\includegraphics{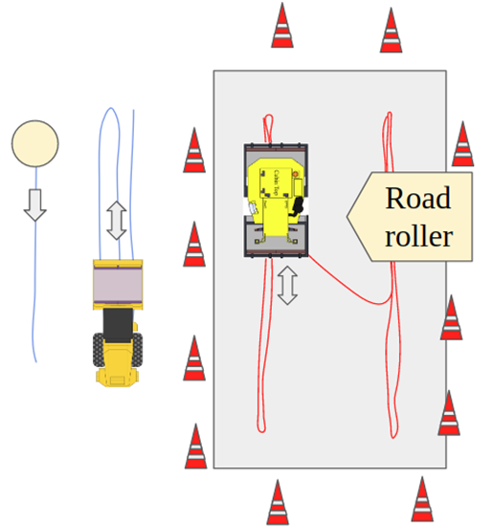}}}\hspace{5pt}
\subfigure[\label{b}Vibration roller]{
\resizebox*{6cm}{!}{\includegraphics{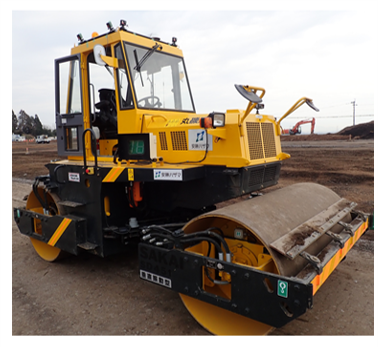}}}
\caption{Experiment setting.}
\label{fig:experimentsite}
\end{center}
\end{figure}

\section{Experimental Evaluations}
\subsection{Testing Environments and Datasets}
To evaluate our proposed approaches, we conducted experiments at two construction sites in Japan with a machine called vibration roller as shown in Figure~\ref{fig:experimentsite}\subref{b}. Vibration roller is used to flatten the earthen basement of structures and facilities. For efficiency of work, there are usually multiple rollers running simultaneously and side by side, thus large moving occlusions become a serious problem for visual SLAM. 
\par In all experiments, a stereo camera was mounted on the cabin top of a roller facing the side. The baseline of the stereo camera was about 1 m. The roller moved along a typical trajectory (Figure~\ref{fig:experimentsite}\subref{a}) with maximum speed of 11 km/h. The ground truth trajectories were recorded using RTK-GNSS. We synchronized ground truth and estimated camera poses by minimizing Absolute Trajectory RMSE (\cite{bescos2018dynaslam, mur2017orbslam2, sturm2012benchmark}) and choosing appropriate time offsets between GNSS's and the camera's timer. Then the estimated camera trajectories will be aligned with ground truth trajectories by Umeyama algorithm \cite{umeyama}. We evaluate the accuracy of camera pose outputs of the vSLAM system with reference to the associated ground truth by Absolute Trajectory RMSE (AT-RMSE).
\par Video data were collected at the site and evaluated in the lab. Image resolution was $3840\times2160$, and frame rate was 60 fps. For efficient evaluation, we downsampled the image sequences to $960\times540$ and 6 fps. We eventually collected five image sequences, three with dynamic machines inside, the 4th one containing only two static machines, and the 5th one was without any occlusions. The specifications of the computer being used were Intel Core i7-9700K CPU, and NVIDIA GeForce RTX 2080 Ti GPU. We used a tool provided by \cite{grupp2017evo} for trajectory accuracy evaluation and visualization.

\par When evaluating our vSLAM system implemetation, all the masks including bounding box and pixel-wise masks are generated beforehand using EfficientDet \cite{tan2019efficientdet} and Detectron2 \cite{wu2019detectron2} version of Mask R-CNN \cite{he2017maskrcnn}. EfficientDet is reported to be able to prioritize detection speed or detection accuracy through configuration. In our implementation, we used EfficientDet-D0 and the weights were trained on MS COCO dataset \cite{lin2014coco}. The weights for Mask R-CNN are also trained on MS COCO datasets \cite{lin2014coco}. Without fine-tuning, they are already good enough for this study. Besides, when calculating overall computation time per frame, we record time consumption for vSLAM tracking part as well as mask generation part respectively, and then add them together. Note that in hierarchical masking, the additional time caused by pixel-wise segmentation will be averaged into all the frames.


\subsection{Performance Evaluation with Our Construction Site Datasets}
\begin{figure}
\begin{center}
\subfigure[\label{a} Absolute position error of every camera pose]{
\resizebox*{7.5cm}{!}{\includegraphics{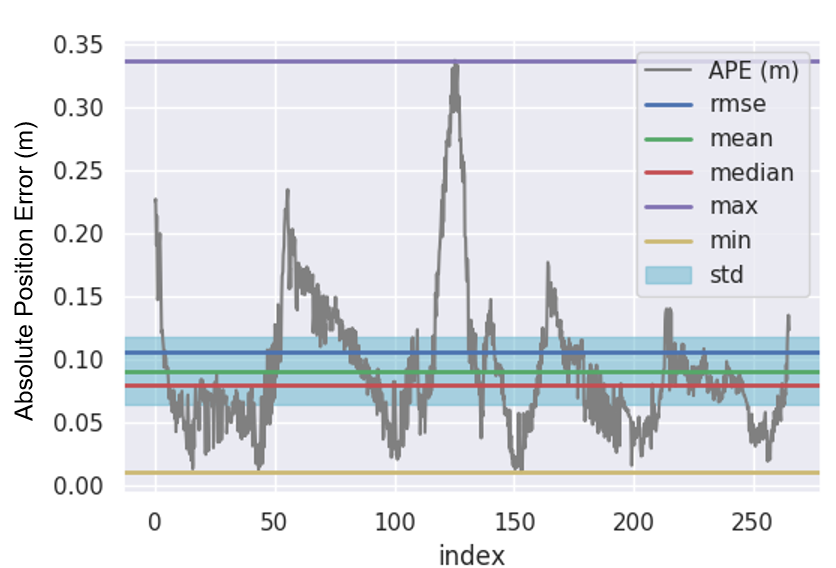}}}\hspace{5pt}
\subfigure[\label{b} Camera trajectory with colormapped position error]{
\resizebox*{7.5cm}{!}{\includegraphics{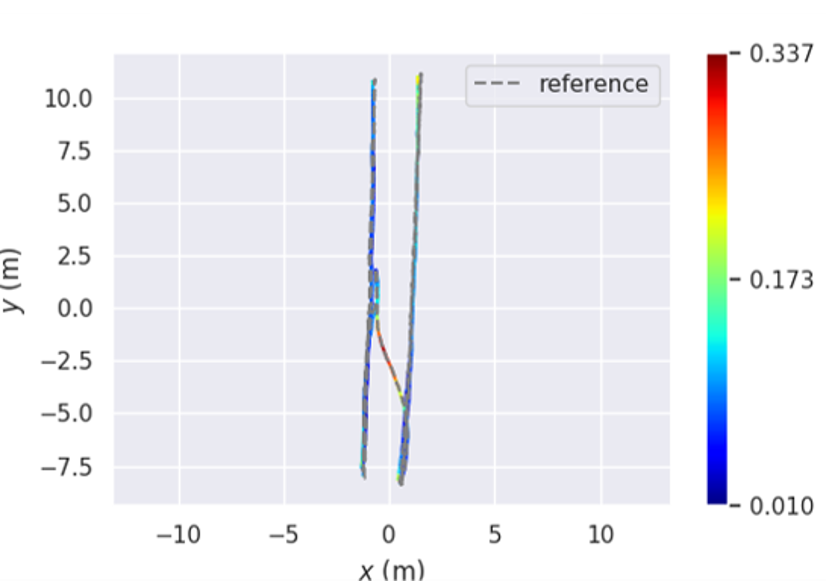}}}
\caption{\label{fig2} Quantitative evaluation for estimated trajectory of image sequence 1 "kumamoto1".}
\label{fig:qeexample}
\end{center}
\end{figure}

\begin{table}[]
\caption{Details about the five image sequences.}
{\begin{tabular}{cccccc} \toprule
\textbf{Dataset details}                                                     & \textbf{kumamoto1}                                                                                                     & \textbf{kumamoto2}                                                                                                     & \textbf{chiba1}                                                                & \textbf{chiba2}    & \textbf{chiba3} \\ \hline
Max. occlusion ratio                                                         & 0.493                                                                                                                  & 0.445                                                                                                                  & 0.521                                                                          & 0.633              & 0.0             \\
MAR\textgreater{}0.5 frames                                                  & 0/1263                                                                                                                 & 0/1186                                                                                                                 & 12/647                                                                         & 69/668             & 0/708           \\
Machines' speed                                                              & 0 to 4 km/h                                                                                                            & 0 to 4 km/h                                                                                                            & 0 to 4 km/h                                                                    & 0 to 4 km/h        & 0 to 4 km/h     \\ \hline
\begin{tabular}[c]{@{}c@{}}Occlusions \& \\ their motion-states\end{tabular} & \begin{tabular}[c]{@{}c@{}}1 roller \\ (dynamic)\\ 7 color cones \\ (static)\\ 1 checkerboard \\ (static)\end{tabular} & \begin{tabular}[c]{@{}c@{}}1 roller \\ (dynamic)\\ 7 color cones \\ (static)\\ 1 checkerboard \\ (static)\end{tabular} & \begin{tabular}[c]{@{}c@{}}1 roller (dynamic)\\ 1 roller (static)\end{tabular} & 2 rollers (static) & no occlusions   \\ 
\bottomrule
\end{tabular}}
\end{table}

\begin{figure}
\begin{center}
\subfigure[\label{a}Estimated trajectory accuracy (lower is better)]{
\resizebox*{7cm}{!}{\includegraphics{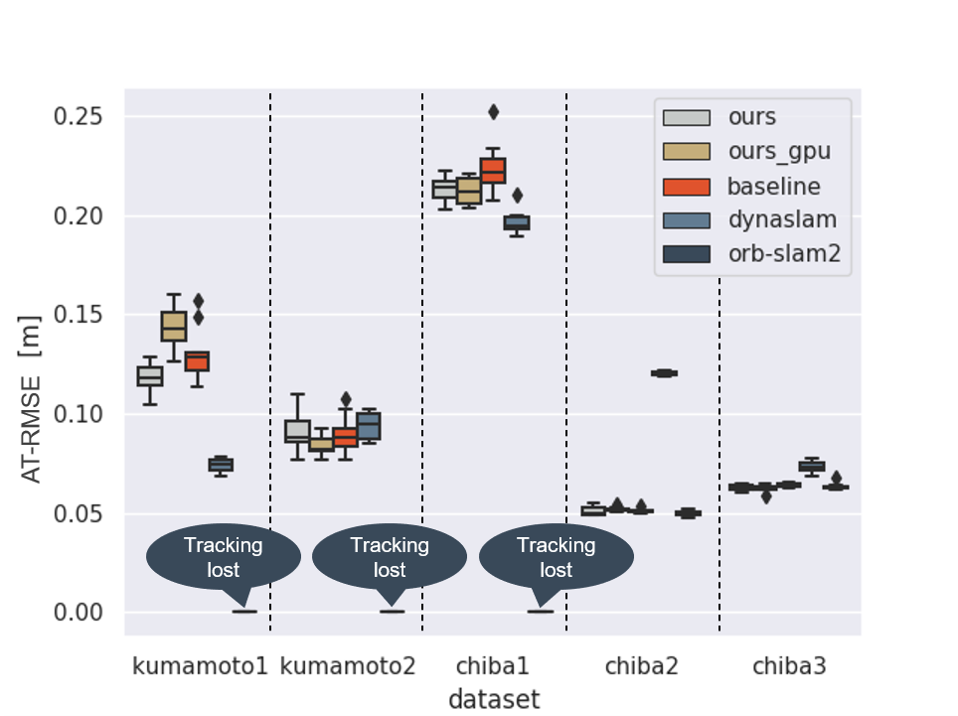}}}\hspace{5pt}
\subfigure[{\label{b}Averaged computation speed (lower is better)}]{
\resizebox*{7cm}{!}{\includegraphics{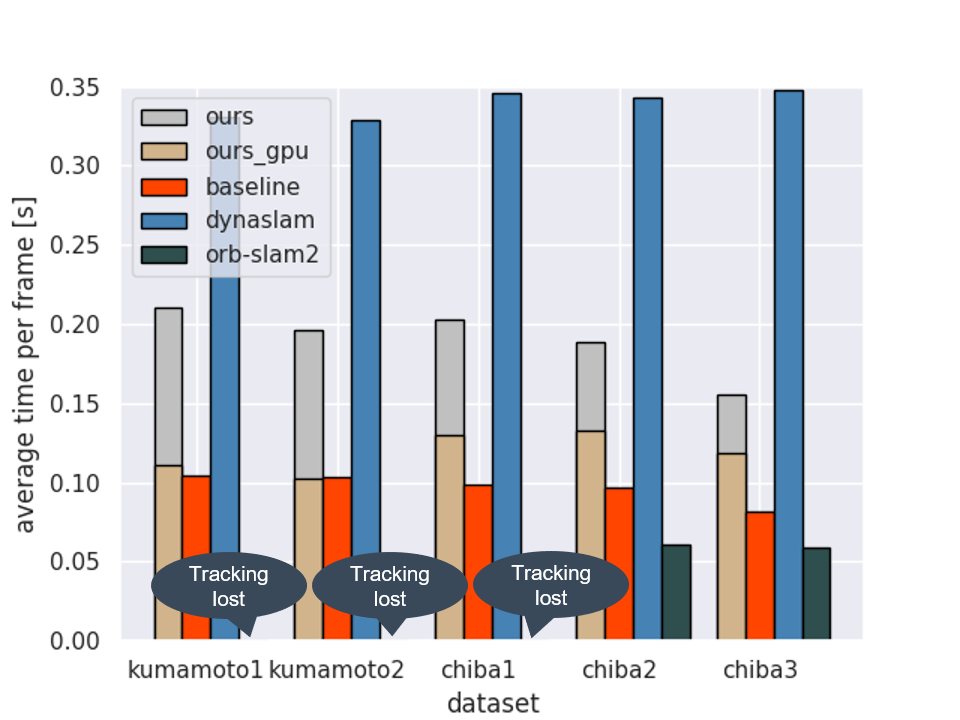}}}
\caption{Performance comparison on our construction site datasets.}
\label{fig:peconstructionsite}
\end{center}
\end{figure}

\par Figure~\ref{fig:qeexample}\subref{a} shows the absolute position error of every camera pose between the estimated trajectory using the proposed system and ground truth of a sequence (kumamoto2). Figure~\ref{fig:qeexample}\subref{b} is a bird’s eye view of the camera trajectory with colormapped absolute position error. There are totally 5 sequences prepared, we repeat such evaluation 10 times for each sequence. The details about the five sequences are described in Table 1. 
Figure~\ref{fig:peconstructionsite}\subref{a} shows the distribution of Absolute Trajectory RMSE of all five sequences. We compare our proposed system with a simple baseline system, with DynaSLAM \cite{bescos2018dynaslam} and with the original ORB-SLAM2 stereo. The baseline system is also based on ORB-SLAM2 but is able to detect and remove moving objects. Its “moving object removal” method is derived from Detect-SLAM \cite{detect-slam2018}, which performs bounding box detection and masks all movable bounding boxes detected. In the results, our proposed system shows better trajectory accuracy in 3 sequences out of five comparing to the baseline, including kumamoto1, chiba1 and chiba3. If the baseline represents fast and efficient handling of dynamic objects, DynaSLAM is much heavier computationally. But the motion segmentation method in DynaSLAM is pixel-level precise and indeed the current state-of-the-art. The experiment results shows that, DynaSLAM does show slight superiority of trajectory accuracy in sequences including kumamoto1, chiba1. The original ORB-SLAM2 stereo can only survive chiba2 and chiba3, which are completely static. In addition, trajectory accuracy of chiba2 and chiba3 are generally better than those of dynamic sequences, no matter which method. Dynamic occlusions do cause irreversible influence on camera ego-motion tracking. 

Averaged computation speed comparisons are shown in Figure~\ref{fig:peconstructionsite}\subref{b}. Our proposed system is relatively slow than the baseline and orb-slam2 stereo at the beginning. However, our method is able to be significantly accelerated by utilizing parallel computing such as GPU acceleration. In our implementation named "ours\_gpu" in Figure~\ref{fig:peconstructionsite}, we enabled GPU acceleration for all the ORB feature extractions, and the speed can be improved notably. However, the trajectory accuracy became different from "ours" to a certain extent, although theoretically they should be the same. We are still looking for the root cause. Finally, time cost of DynaSLAM (only tracking, without background inpainting) is 2 to 3 times of ours\_gpu. Large computation latency is not preferable, since our targeted task is real-time positioning and navigation of a construction machine.

\subsection{Ablation Study}
\subsubsection{\textit{Hierarchical Object Masking}}

\begin{figure}
\begin{center}
\subfigure[\label{a}Three machines working parallelly to each other.]{
\resizebox*{10cm}{!}{\includegraphics{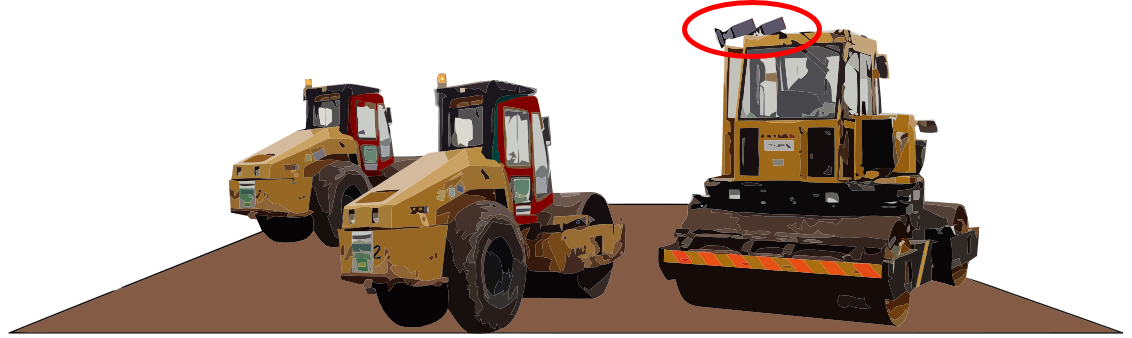}}}\hspace{5pt}
\subfigure[\label{b}From view point of the on-board camera]{
\resizebox*{5.5cm}{!}{\includegraphics{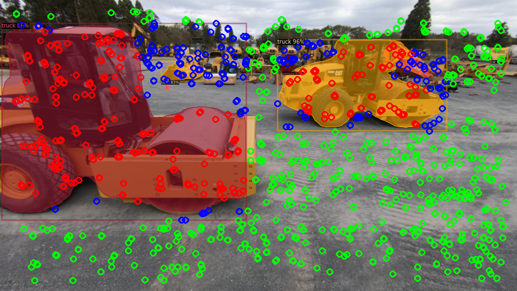}}}
\caption{\label{fig2} Dynamic scene and hierarchical masking example.}
\label{fig:hmaskeval}
\end{center}
\end{figure}

\begin{figure}[t!]
    \centering
    \includegraphics[scale=0.35]{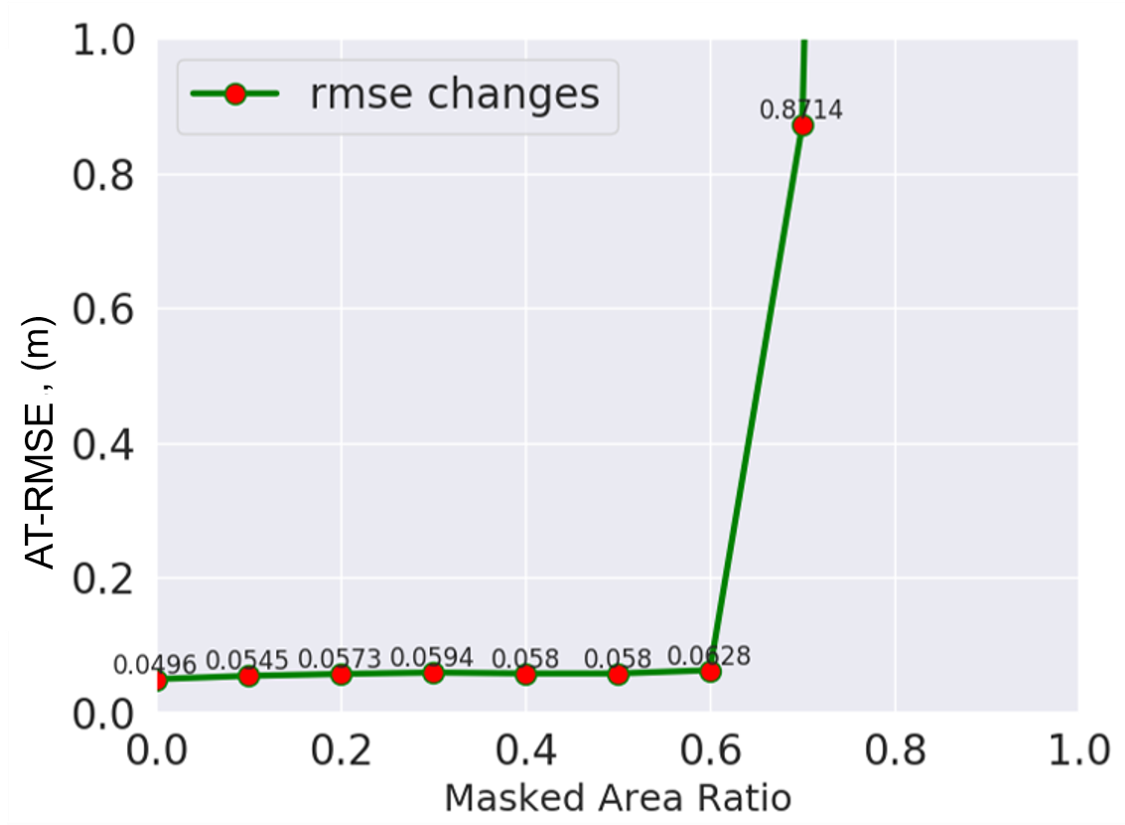}
    \caption{A toy experiment: estimated trajectory accuracy when putting different sizes of occlusions on the 4th image sequence "chiba2".}
    \label{fig:toyexperiment}
\end{figure}

Hierarchical masking aims to efficiently propose an appropriate initial mask in case there are overly-large dynamic occlusions in the image. Figure~\ref{fig:hmaskeval}\subref{a} shows a scene when the machine was working along with two other machines and thus had two large occlusions in the camera view. Figure~\ref{fig:hmaskeval}\subref{b} shows a sample image recorded by the on-board camera. Notice that the two rectangles labeled as truck are bounding boxes detected by object detection algorithm, and the color masks inside the bounding boxes are by pixel-wise segmentation. Besides, ORB feature points are extracted and marked on this image. Green points are static features on the static background, blue points are those included by bounding boxes but not included by pixel-wise masks, and red points are features masked by pixel-wise masks. It is obvious that bounding box mask causes many innocent static features being treated as dynamic. Through a toy experiment, we can see how it will cause shortage of available feature points and lead to worse pose tracking accuracy. Then by a real example in our datasets, we explain the effectiveness of hierarchical masking.\\
\newpage
\noindent (1) \ \textit{A toy experiment}\\
We put a fake constant dynamic occlusion at the center of the mask images of the 4th image sequence chiba2 (static scene). And we adjusted the size of this area to simulate different occlusion ratio and see how the result trajectory accuracy changes. The result is plotted in Figure~\ref{fig:toyexperiment}. Before occlusion ratio reaches 0.6, trajectory error only varies over a small range; when occlusion ratio exceeds 0.7, the RMSE increases exponentially due to shortage of available features. Therefore, when occlusion ratio of the image approaches the critical point of 0.6, we define it as a large occlusion condition, requiring the refinement of the bounding box mask to a pixel-wise mask to suppress the growing error. Besides, when occlusion ratio is larger than 0.6, tracking lost will frequently happen which is not preferred when navigating a construction machine. To avoid tracking lost and relocalization, we set the threshold ($\tau_{\rm mar}$ in section~3.3) to 0.5 as a safty limit. 

However, when occlusion ratio is far smaller than 0.6, bounding box mask is enough and also faster to get. With our computer, generating bounding box masks for one image frame takes 0.0207 seconds in average while a pixel-wise mask takes 0.12 seconds.\\

\noindent (2) \ \textit{An overly large occlusion case}
\par In order to demonstrate the effectiveness of hierarchial masking when facing overly large occlusions, we show an example in sequence "chiba2". From the 3500th frame to 4500th frame (1000 frames in the original 60 fps sequence) in "chiba2" sequence, we encountered an overly large occlusion. As Table~2 shows, when changing from bounding box mask to pixel-wise mask, the maximum masked area ratio reduced from 0.63 to 0.32 and, correspondingly, trajectory error decreased. Hierarchical masking benefits trajectory accuracy, and it will cost much less time than only using pixel-wise mask. In this example, only 2/3 of the frames during this period need to calculate pixel-wise mask. And the maximum masked area ratio is constrained within 0.5. Note that although the Absolute Trajectory RMSE difference between 0.0404 and 0.0437 seems trivial here in Table~2. It is partially because of the trajectory alignment algorithm \cite{umeyama} we used for evaluation, the actual accuracy difference can be larger. 

\begin{table}
\caption{Tracking accuracy of "chiba2" with three different mask types.}
\centering
{\begin{tabular}{lllll} \toprule
\textbf{Mask type}   & \textbf{\begin{tabular}[c]{@{}l@{}}AT-RMSE, m\\ (average of 10 trials)\end{tabular}} & \textbf{\begin{tabular}[c]{@{}l@{}}Max. occlusion ratio\end{tabular}} &  &  \\ \cline{1-3} \\
B-box mask           &     \qquad 0.0437                                                                                                 &  \quad 0.63                                                                       &  &  \\
Hierarchical mask &    \qquad 0.0404                                                                                                 &  \quad 0.50                                                                       &  &  \\
Pixel-wise mask      &    \qquad 0.0397                                                                                                 &  \quad 0.32                                                                       &  &  \\
\bottomrule
\end{tabular}}
\end{table}

\begin{figure}[t!]
    \centering
    \includegraphics[scale=0.58]{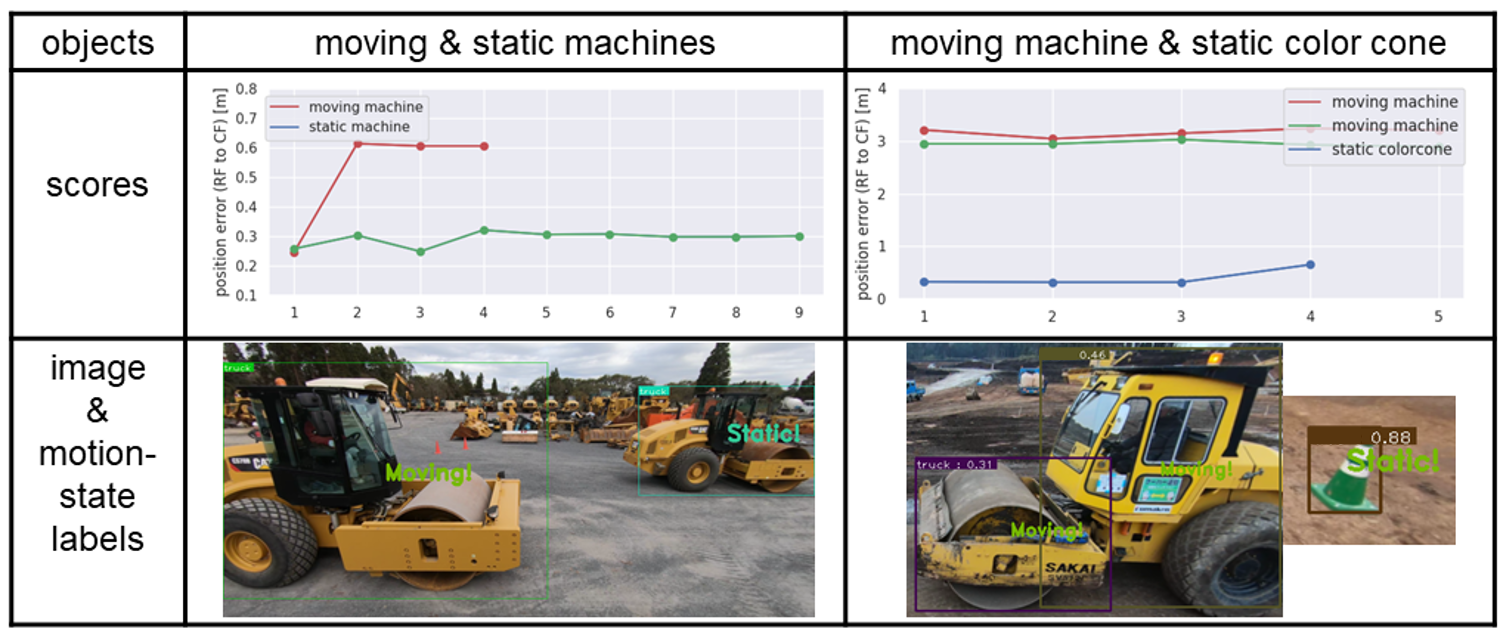}
    \caption{Illustration of the classification result. In the left column, the third row shows that there is one machine classified as "moving" and another classified as "static" in this frame. The second row shows the position errors of 3D points on these two machines between this frame and the reference frame. Points on the "moving" machine have higher position errors. Similarly, in the right column, there are also "moving" machines (two parts of one machine) and a "static" color cone detected.}
    \label{fig:SDdist}
\end{figure}

\subsubsection{\textit{Objects' Motion-state Classification}}
Not all a priori dynamic objects are moving. Ignoring static objects leads to loss of information, especially when they are salient and occupy a large area in the image. Therefore, we designed the objects' motion-state classification algorithm to detect static objects and unmask them for ego-motion tracking. Figure~\ref{fig:SDdist} shows dynamic and static objects detected in the image sequences and scores relating to the possibility of them being dynamic. We also show an example of using the proposed algorithm in visual SLAM. Again, we use the 3500th frame to 4500th frame (1000 frames) in "chiba2" sequence, and since the machines are totally static during this period, they are detected as static and unmasked. Table~3 shows how it can influence the tracking accuracy.

\begin{table}
\centering
\caption{Tracking accuracy with motion-state classification.}
{\begin{tabular}[l]{@{}lcc} \toprule
  \textbf{Mask type}
  & \textbf{AT-RMSE, m} & \textbf{Max. occlusion ratio} \\ \cline{1-3} \\
  All objects masked & 0.04973 & 0.63 \\
  Static objects unmasked & 0.04198 & 0.0 \\
\bottomrule
\end{tabular}}
\end{table}

\par
\begin{figure}[t!]
    \centering
    \includegraphics[scale=0.45]{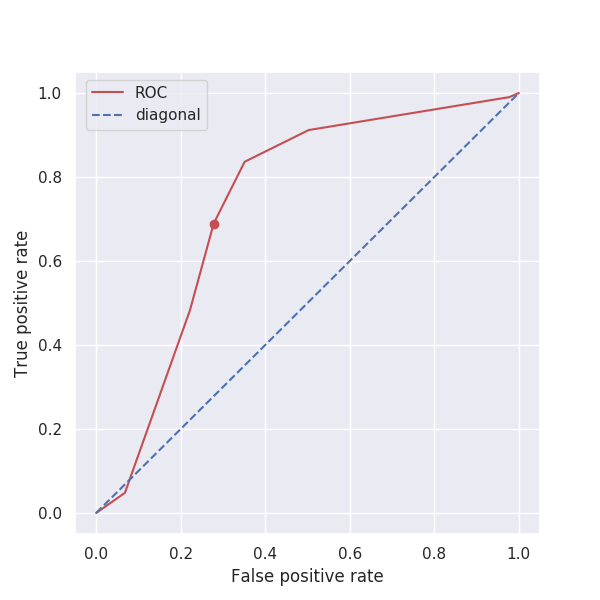}
    \caption{ROC curve for the motion-state classification when $\sigma_{\mathrm{bkg}}$ was between 0 and 0.6, estimated with the 3rd image sequence "chiba1". The Area Under Curve (AUC) reached 0.737. Red dot is the position where $\sigma_{\mathrm{bkg}}=0.12$.}
    \label{fig:roc}
\end{figure}

 However, there is still one bottleneck in this algorithm. $\sigma_{bkg}$ is an essential parameter for the performance of motion-state classification. For all the evaluations above with the four image sequences, we set $\sigma_{bkg}$ to 0.12 which was empirically determined. To explore the influence of this parameter on system performance, we variate $\sigma_{bkg}$ between 0 and 0.6 to evaluate the classifier in terms of ROC (Receiver Operating Characteristics). Since the final target is to find static objects, "static" is regarded as positive and "dynamic" as negative, ignoring objects that cannot be classified. The ROC curve is shown in Figure~\ref{fig:roc}, true positive rate (TPR, sensitivity) on the y axis is the ratio of true positive number over the sum of true positives and false negatives. False positive rate on the x axis is the ratio of false positives over the sum of false positives and true negatives. According to this curve, the Area Under the Curve (AUC) reached 0.737, which proved it to be a valid classifier. The red dot in the plot is the position where $\sigma_{bkg}=0.12$.

\subsection{Evaluation with KITTI Dataset}
The KITTI Dataset \cite{geiger2013vision} provides stereo camera sequences in outdoor urban and highway environments. It has been a wide-spread benchmark for evaluating vSLAM system performance, especially trajectory accuracy. Works such as \cite{bescos2018dynaslam, mur2017orbslam2} all provide evaluation results with KITTI. There are some sequences in KITTI containing normal-size dynamic occlusions, thus KITTI is also appropriate for evaluation of our method. Table 4 shows the evaluation results. The comparison includes four systems, our proposed system, the baseline, DynaSLAM and ORB-SLAM2 stereo, same as in section 4.2. For the baseline, DynaSLAM and ORB-SLAM2 stereo, all the settings remain the same as before. For our system, $\tau_{\rm mar}$ (Section~3.3) remains to be 0.5 and $\sigma_{\rm bkg}$ (Section~3.4) remains to be 0.12. However, $N$ (Section~3.4) is changed to be 2, since frame rate of KITTI is 10 fps and the cars are much faster than our construction machines. We ran each sequence 10 times with each system and recorded the averaged Absolute Trajectory RMSE (AT-RMSE, m) as well as the averaged computation time per frame (s). For our system, we recorded both results with GPU acceleration (w A) and without GPU acceleration (w/o A). Between the four comparisons, best AT-RMSE for each sequence is marked with bold font and best computation time marked with bold, italic font. Note that the AT-RMSE results of DynaSLAM and ORB-SLAM2 stereo are different from the original paper. It is because we only align the trajectory with ground truth without adjusting scale before calculating trajectory error, since our target is online positioning with vSLAM.

From Table~4, we see that in terms of computation speed, ORB-SLAM2 stereo is always the best. Because it has adapted the static environment assumption. DynaSLAM is the slowest. Ours is slightly worse than the baseline and ORB-SLAM2 stereo, however, we do see that GPU acceleration helps improving speed to a tolerable level. In terms of AT-RMSE, the results are various, but DynaSLAM and ORB-SLAM2 stereo did have the most bold fonts numbers. In KITTI dataset, there are moving automobiles, bicycles and persons in some frames, but they are not overly-large. Actually there are only 6 frames in "07" in which occlusion ratio became larger than 0.5. Besides, automobiles on the street do not contain so many salient feature points as construction machines, most of them have texture-less and smooth surface. Therefore, our proposed system is not advantageous in KITTI.

\begin{table}
\footnotesize
\caption{Trajectory accuracy and time consumption evaluation on KITTI Dataset.}
{\begin{tabular}{lllllllllll} 
\toprule
\hline
\multicolumn{1}{l|}{\multirow{3}{*}{\textbf{Sequence}}} & \multicolumn{4}{c|}{\textbf{ours}}                                                                                                                                    & \multicolumn{2}{c|}{\textbf{baseline}}                                                                                                                                & \multicolumn{2}{c|}{\textbf{dynaslam (tracking)}}                                                                                                                                & \multicolumn{2}{c|}{\textbf{orb-slam2}}                                                                                                          \\ \cline{2-11} 
\multicolumn{1}{l|}{}                                   & \multicolumn{2}{l|}{{\fontsize{7.5}{7.2}\textbf{AT-RMSE (m)}}}                                   & \multicolumn{2}{l|}{\textbf{\begin{tabular}[c]{@{}l@{}}time per \\ frame (s)\end{tabular}}} & \multicolumn{1}{l|}{\multirow{2}{*}{\textbf{\begin{tabular}[c]{@{}l@{}}\fontsize{7.5}{7.2}\textbf{AT-RMSE} \\ \fontsize{7.5}{7.2}\textbf{(m)}\end{tabular}}}} & \multicolumn{1}{l|}{\multirow{2}{*}{\textbf{\begin{tabular}[c]{@{}l@{}}time per \\ frame (s)\end{tabular}}}} & \multicolumn{1}{l|}{\multirow{2}{*}{\textbf{\begin{tabular}[c]{@{}l@{}}\fontsize{7.5}{7.2}\textbf{AT-RMSE} \\ \fontsize{7.5}{7.2}\textbf{(m)}\end{tabular}}}} & \multicolumn{1}{l|}{\multirow{2}{*}{\textbf{\begin{tabular}[c]{@{}l@{}}time per \\ frame (s)\end{tabular}}}} & \multicolumn{1}{l|}{\multirow{2}{*}{\textbf{\begin{tabular}[c]{@{}l@{}}\fontsize{7.5}{7.2}\textbf{AT-RMSE} \\ \fontsize{7.5}{7.2}\textbf{(m)}\end{tabular}}}} & \multirow{2}{*}{\textbf{\begin{tabular}[c]{@{}l@{}}time per \\ frame (s)\end{tabular}}} \\ \cline{2-5}
\multicolumn{1}{l|}{}                                   & \multicolumn{1}{l|}{\textbf{w/o A}} & \multicolumn{1}{l|}{\textbf{w A}} & \multicolumn{1}{l|}{\textbf{w/o A}}           & \multicolumn{1}{l|}{\textbf{w A}}           & \multicolumn{1}{l|}{}                                  & \multicolumn{1}{l|}{}                                                                                        & \multicolumn{1}{l|}{}                                  & \multicolumn{1}{l|}{}                                                                                        & \multicolumn{1}{l|}{}                                  &                                                                                         \\ \hline
KITTI 00                                                & 2.1290                              & \textbf{1.7304}                   & 0.2018                                        & 0.1565                                      & 2.0173                                                 & 0.0912                                                                                                       & 3.9691                                                 & 0.3354                                                                                                       & \textbf{1.7304}                                        & \textit{\textbf{0.0703}}                                                                \\
KITTI 01                                                & \textbf{8.4940}                     & 8.7620                            & 0.1860                                        & 0.1305                                      & 9.1271                                                 & 0.0917                                                                                                       & 21.8982                                                & 0.3273                                                                                                       & 8.7620                                                 & \textit{\textbf{0.0734}}                                                                \\
KITTI 02                                                & 5.1759                              & \textbf{4.7338}                   & 0.1764                                        & 0.1194                                      & 4.9280                                                 & 0.0935                                                                                                       & 5.9401                                                 & 0.3243                                                                                                       & 4.9994                                                 & \textit{\textbf{0.0771}}                                                                \\
KITTI 03                                                & 3.2169                              & 3.4246                            & 0.1462                                        & 0.0983                                      & 3.1174                                                 & 0.0898                                                                                                       & 4.7770                                                 & 0.3459                                                                                                       & \textbf{3.0735}                                        & \textit{\textbf{0.0723}}                                                                \\
KITTI 04                                                & 1.0835                              & 1.2937                            & 0.1811                                        & 0.1297                                      & \textbf{0.9970}                                        & 0.0864                                                                                                       & 1.3371                                                 & 0.3420                                                                                                       & 1.0079                                                 & \textit{\textbf{0.0672}}                                                                \\
KITTI 05                                                & 2.1243                              & 2.2529                            & 0.1915                                        & 0.1382                                      & 2.0528                                                 & 0.0923                                                                                                       & \textbf{1.7644}                                        & 0.3482                                                                                                       & 1.9751                                                 & \textit{\textbf{0.0717}}                                                                \\
KITTI 06                                                & 2.1718                              & 2.2651                            & 0.2076                                        & 0.1546                                      & 1.9338                                                 & 0.0943                                                                                                       & 2.0627                                                 & 0.3434                                                                                                       & \textbf{1.8793}                                        & \textit{\textbf{0.0752}}                                                                \\
KITTI 07                                                & 1.2323                              & 1.3159                            & 0.1791                                        & 0.1337                                      & 1.1799                                                 & 0.0843                                                                                                       & 1.1285                                                 & 0.3493                                                                                                       & \textbf{0.9733}                                        & \textit{\textbf{0.0632}}                                                                \\
KITTI 08                                                & 4.5641                              & 5.2294                            & 0.1945                                        & 0.1445                                      & 4.7857                                                 & 0.0882                                                                                                       & \textbf{3.7062}                                        & 0.3488                                                                                                       & 4.6483                                                 & \textit{\textbf{0.0675}}                                                                \\
KITTI 09                                                & 4.9692                              & 5.8698                            & 0.1760                                        & 0.1231                                      & 7.1441                                                 & 0.0865                                                                                                       & \textbf{4.2753}                                        & 0.3463                                                                                                       & 5.9788                                                 & \textit{\textbf{0.0657}}                                                                \\
KITTI 10                                                & 2.5849                              & 2.6375                            & 0.1522                                        & 0.1022                                      & 2.6986                                                 & 0.0912                                                                                                       & \textbf{2.2028}                                        & 0.3466                                                                                                       & 2.6699                                                 & \textit{\textbf{0.0631}}                                                                \\ \hline
\bottomrule
\end{tabular}}
\end{table}

\section{Conclusions \& Future Work}
We presented a stereo vSLAM system for dynamic outdoor construction sites. The key contributions are, first, a hierarchical masking strategy that can timely refine overly-large occlusion mask in an efficient way. Second, a semantic-geometric approach for objects' motion-state classification and a two-step coarse-to-fine ego-motion tracking scheme. Our system accurately retrieved the motion trajectories of a stereo camera at construction sites, and most of the surrounding objects' motion-states in the scene were correctly predicted. Hierarchical object masking has also been proved to be a simple but useful strategy. Our proposed system can deal with dynamic and crowded environments that standard vSLAM systems may fail to keep tracking.
\par In future work, the method to select reference frames can be optimized to enable more robust object motion-state classification. Moreover, we plan to combine vSLAM with an inertial measuring unit (IMU) sensor for higher-accuracy positioning. However, the fierce and high-frequency vibration of the vibration roller may cause severe noises in the IMU measurements, which results in worse pose accuracy. Therefore, we will look into this problem and meanwhile also explore other topics about visual SLAM related research at construction sites.

\label{lastpage}


\end{document}